\documentclass[letterpaper]{article} % DO NOT CHANGE THIS

\usepackage{aaai2027} % DO NOT CHANGE THIS

\usepackage[T1]{fontenc}
\usepackage[utf8]{inputenc}

\usepackage{algorithm}
\usepackage{algpseudocode}
\usepackage{amsmath}
\usepackage{amssymb}

\usepackage{xcolor}
\usepackage{graphicx} % DO NOT CHANGE THIS
\usepackage{booktabs}
\usepackage{tabularx}

\usepackage[hyphens]{url} % DO NOT CHANGE THIS
\usepackage{natbib}  % DO NOT CHANGE THIS
\usepackage{caption} % DO NOT CHANGE THIS
\usepackage{subcaption}

\usepackage{microtype}

\usepackage{newfloat}
\usepackage{listings}
\usepackage[most]{tcolorbox}

\DeclareCaptionStyle{ruled}{
  labelfont=normalfont,
  labelsep=colon,
  strut=off
} % DO NOT CHANGE THIS

\definecolor{RecordFrame}{HTML}{666666}
\definecolor{RecordBack}{HTML}{F4F4F4}
\definecolor{RecordTitle}{HTML}{666666}

\lstdefinestyle{recordjson}{
  basicstyle=\ttfamily\fontsize{6.0}{7.1}\selectfont,
  columns=fullflexible,
  keepspaces=true,
  showspaces=false,
  showtabs=false,
  showstringspaces=false,
  tabsize=2,
  breaklines=true,
  breakatwhitespace=true,
  breakautoindent=true,
  breakindent=1.5em,
  numbers=none,
  numbersep=0pt,
  xleftmargin=0pt,
  framexleftmargin=0pt,
  xrightmargin=0pt,
  aboveskip=0pt,
  belowskip=0pt,
  upquote=true,
  literate={…}{{\ensuremath{\ldots}}}1
}

\tcbset{
  recordbox/.style={
    enhanced,
    sharp corners,
    arc=0pt,
    outer arc=0pt,
    colframe=RecordFrame,
    colback=RecordBack,
    colbacktitle=RecordTitle,
    coltitle=white,
    boxrule=0.55pt,
    left=1.45mm,
    right=1.45mm,
    top=1.00mm,
    bottom=1.00mm,
    before skip=0pt,
    after skip=0pt,
    fonttitle=\sffamily\bfseries\fontsize{7.1}{8.1}\selectfont,
    lefttitle=1.45mm,
    righttitle=1.45mm,
    toptitle=0.60mm,
    bottomtitle=0.60mm,
    titlerule=0pt,
    listing engine=listings,
    listing only,
    listing options={
      style=recordjson,
      numbers=none,
      numbersep=0pt,
      xleftmargin=0pt,
      framexleftmargin=0pt
    }
  },
  instructionbox/.style={recordbox},
  inputbox/.style={recordbox},
  outputbox/.style={recordbox},
  chosenbox/.style={recordbox},
  rejectedbox/.style={recordbox},
  promptbox/.style={recordbox},
  metadatabox/.style={recordbox}
}

\title{Large Language Model for Operations Research Formulation Selection in Multi-Warehouse Inventory Allocation}
\author{
    Jintao Xu\equalcontrib,
    Yingzheng Ma\equalcontrib,
    Jiong Dong\equalcontrib,
    Yongzhi Qi,\thanks{Corresponding authors.},
    Jianshen Zhang\footnotemark[2]
}
\affiliations{
     Supply Chain Tech Team Y, JD.com, Beijing, China\\
    \{xujintao.3014,  mayingzheng.1, dongjiong.1, qiyongzhi1, zhangjianshen\}@jd.com
}

\begin{document}

\nocopyright
\maketitle

\begin{abstract}
Multi-warehouse inventory allocation is typically formulated as a mixed-integer programming (MIP) problem, yet no single formulation consistently matches heterogeneous instance-level regimes induced by demand concentration, inventory imbalance, replenishment scale, service constraints, and forecast volatility. 
We study this issue as instance-wise operations research (OR) formulation selection, where each allocation instance is assigned to a solver-executable formulation from a candidate OR expert library.
We propose a solver-guided large language model (LLM) framework for OR formulation selection, in which each OR expert corresponds to a MIP formulation encoding a distinct allocation priority. To train the selector, the framework first constructs balanced expert-conditioned supervised fine-tuning (SFT) records for schema learning, and then uses MIP solver evaluation on historical instances to convert solver-evaluated allocation-quality gaps into margin-weighted identity preference optimization (IPO) preferences and per-instance expert-score metadata for reward lookup during group relative policy optimization (GRPO) to assign rewards to sampled responses.
Experiments on multi-warehouse inventory allocation instances from JD$\mathord{.}$com, one of China's largest e-retailers, demonstrate that GRPO substantially improves expert-selection accuracy relative to the SFT+IPO selector and, more importantly, produces higher realized allocation quality than both the preference-trained selector and the best fixed formulation.
With GRPO, Hit Ratio@1 and Hit Ratio@2 increase from 21.45\% to 50.42\% and from 70.47\% to 82.31\%. The resulting selector achieves an allocation accuracy gain of 12.57 percentage points over the incumbent baseline, outperforming both the SFT+IPO selector and the best fixed OR expert, and reduces the gap to the ex-post oracle to 4.85 percentage points.

\end{abstract}

\section{Introduction}

Multi-warehouse inventory allocation is a fundamental decision problem in retail supply-chain operations. Given a stock-keeping unit (SKU) and a fixed replenishment quantity, the task is to allocate integer replenishment volumes across warehouses according to current inventory, forecast demand, and additional constraints. A common objective is to balance warehouse-level target inventory days (TID) around a system-wide coverage target, so that under-covered warehouses receive replenishment while warehouses with excessive inventory are not further overstocked. This objective naturally leads to operations research (OR) formulations that preserve total replenishment and optimize allocation quality under warehouse-level constraints.

In practical allocation systems, however, TID balancing is not well captured by a single fixed formulation. Different instances may emphasize different structural regimes, such as satisfying admissible TID bands, minimizing aggregate deviation from the system-wide TID, preserving lexicographic service priorities, or accounting for heterogeneous forecast reliability. These regimes are shaped by interactions among demand concentration, replenishment scale, warehouse-level inventory states, lower and upper allocation bounds, selective service requirements, and forecast volatility. Consequently, a formulation that is appropriate for one SKU instance may be suboptimal or structurally mismatched for another, even when both instances belong to the same business scenario.

This motivates a learning-based selection problem that we call instance-wise OR formulation selection. 
Rather than applying a fixed formulation to all instances, the system selects one formulation from a candidate library for each allocation instance. The selection depends on interactions among SKU-level demand patterns, warehouse-level inventory states, constraints, and the allocation quality realized after optimization.
Thus, effective formulation selection requires learning from posterior solver evidence rather than relying only on pre-specified regime heuristics.

We propose a solver-guided large language model (LLM) framework for instance-wise operations research (OR) formulation selection.
Given an allocation instance and a library of candidate OR experts, the framework learns a high-level mapping from heterogeneous instance characteristics to an appropriate OR expert. The selected
expert then instantiates a MIP formulation, which
is optimized by an MIP solver to obtain the
final allocation plan. This division of labor combines
language-model-based instance understanding with the reliability and executability of explicit OR formulations.

We instantiate the candidate expert library with four complementary mixed-integer programming (MIP) formulations: lexicographic band (LB), scalarized band-deviation (SBD) \citep{xu2026orla}, deviation minimization (DM) \citep{xu2026orla}, and reliability-calibrated band (RCB). These formulations share the same replenishment-conservation structure, but encode different optimization priorities. LB first maximizes the replenishment volume allocated within the target TID band and then minimizes residual deviation. SBD jointly trades off band satisfaction and deviation through a weighted objective. DM directly minimizes aggregate warehouse-level deviation from the all-warehouse TID. RCB incorporates forecast variability by imposing larger penalties on band violations at warehouses with more reliable demand signals. This formulation library provides an explicit and interpretable candidate space for adaptive allocation.

We develop a progressive solver-guided post-training procedure that aligns the LLM selector with formulation quality.
For each historical instance, we solve all candidate formulations and evaluate their resulting allocation plans, producing instance-formulation scores as ex-post optimization supervision. These scores are then converted into three levels of training signals. First, supervised fine-tuning (SFT) \citep{ouyang2022training} trains the language model to produce schema-consistent formulation-selection outputs. Second, identity preference optimization (IPO) \citep{GheshlaghiAzar2024} constructs margin-weighted chosen-rejected pairs from within-instance score gaps, encouraging the model to distinguish formulations that differ in realized allocation quality. Third, group relative policy optimization (GRPO) \citep{shao2024deepseekmath,deepseekai2025r1} further refines the selector by comparing multiple sampled formulation choices for the same instance using metadata and rewards.

Empirically, we evaluate the proposed selector on real-world multi-warehouse inventory allocation instances from  JD.com, one of China’s largest e-retailers. We compare the full SFT+IPO+GRPO selector against the best fixed OR expert, an SFT+IPO selector, and the ex-post oracle, assessing both expert-selection accuracy and realized allocation quality. 
Expert-selection accuracy is measured by Hit Ratio@1 (HR@1) and Hit Ratio@2 (HR@2). For each instance, we first solve all four OR experts and rank them by their allocation accuracy. HR@1 indicates whether the single expert selected by the LLM is the top-ranked expert, while HR@2 indicates whether the selected expert falls within the top two experts in this ranking.
The results show that GRPO substantially improves formulation selection over SFT+IPO on the observed instance distribution, increasing HR@1 by 28.97 percentage points (pp) and HR@2 by 11.84 pp. This improvement further translates into higher allocation utility: the SFT+IPO+GRPO selector improves allocation accuracy by 12.57 pp over the incumbent baseline, outperforming both the SFT+IPO selector and the best fixed expert while reducing the gap to the oracle. Further oracle-conditioned analysis shows that the gains are concentrated in the dominant expert regime.

\subsection{Contributions}
Our main contributions are summarized as follows:

\begin{itemize}
    \item We formulate heterogeneous multi-warehouse inventory
    allocation as an instance-wise OR formulation selection problem
    and introduce a solver-guided LLM framework that routes each
    allocation instance to a MIP expert.

    \item We construct a complementary library of four OR experts
    and develop a progressive solver-guided post-training pipeline
    combining balanced SFT, margin-weighted IPO, and GRPO.
    For each historical instance, all candidate experts are evaluated
    offline by a MIP solver. The resulting score gaps are used to
    construct IPO preference pairs, while expert-wise scores and
    rankings are cached as per-instance GRPO reward metadata.

    \item We evaluate the resulting formulation selector on real-world
    multi-warehouse inventory allocation instances from JD.com. The results show that GRPO substantially improves expert-selection accuracy over the SFT+IPO selector and yields the best realized allocation quality.

\end{itemize}

\section{Related Work}\label{sec:related-work}
\subsection{LLM-Based Routing and Selection}
A growing body of work studies LLMs and learned policies as high-level selectors over predefined decision modules, including experts, operators, solvers, algorithms, and reasoning paradigms. These studies share the common view that complex problem instances often exhibit heterogeneous structures, and that no single expert or algorithm is uniformly optimal across all instances.

One closely related direction uses LLMs to select among specialized experts or operators. For example, \cite{du2024moe_network_opt} employ an LLM-enabled gating mechanism to analyze objectives and constraints, select specialized DRL experts, and aggregate their decisions for wireless network optimization. Similar ideas have also appeared in evolutionary optimization. LAOS \cite{zhang2025laos} uses an LLM for adaptive operator selection, while PAIR \cite{ali2025pair} leverages an LLM to guide pair selection during evolutionary search. These methods demonstrate that LLMs can serve as high-level controllers for choosing among predefined optimization components.

Another related line of work focuses on learned solver or algorithm selection. Such methods exploit the complementary strengths of different solvers or algorithms across problem instances. Representative approaches include feature-based neural solver selection \cite{gao2025neural}, as well as LLM-based algorithm selection through problem--algorithm representation matching \cite{wu2024llm_algorithm_selection}. Compared with expert or operator selection, this line of work emphasizes instance-level performance prediction over candidate solvers or algorithms, which is closely related to the broader algorithm selection and algorithm portfolio literature.

Recent studies further extend the selection idea to model and reasoning-paradigm routing. Select-then-Solve \cite{zhou2026select} learns to select suitable reasoning paradigms for different tasks, while Router-R1 \cite{zhang2025routerr1} investigates RL-based routing among multiple LLMs. These works suggest that routing is not limited to selecting optimization solvers or operators, but can also be used to choose among heterogeneous reasoning strategies or model experts. This paradigm is particularly relevant to multi-warehouse inventory allocation, where heterogeneous operational constraints require adaptive expert selection and formulation choice.

\subsection{LLMs for Operations Research}
Beyond general routing and expert selection, LLMs have recently been explored as modeling and solving assistants for OR, where they translate natural-language into OR formulation and executable
code calling solver API \cite{Astorga2024,Jiang2024,Huang2025}. 
In multi-warehouse inventory allocation, \cite{xu2026orla} proposed an LLM method that generates and selects OR formulations for allocation tasks under solver-side feasibility checking. Our work follows the solver-grounded perspective, but addresses a different problem setting. Rather than asking the LLM to generate a new formulation or related code for each instance, our approach trains an LLM-based selector to route each allocation instance to an explicit candidate OR formulation.

\subsection{Post-Training Methods for LLMs}

Our post-training pipeline is most closely related to three components: Supervised Fine-Tuning (SFT), preference optimization, and reinforcement learning (RL). SFT is widely used as the initial alignment stage for adapting pretrained language models to instruction-following behavior \citep{ouyang2022training}. Building on such initialization, recent work has increasingly explored offline preference optimization. In particular, Identity Preference Optimization (IPO) provides a direct objective for learning from pairwise preferences without requiring explicit reward-model training or online reinforcement learning \citep{GheshlaghiAzar2024}. For further policy improvement, RL-based post-training remains an important direction. While Proximal Policy Optimization (PPO) \citep{schulman2017ppo} is a standard baseline, recent reasoning-oriented studies have explored more efficient variants such as Group Relative Policy Optimization (GRPO) \citep{shao2024deepseekmath,deepseekai2025r1}, which derives training signals from within-group relative reward comparisons rather than an explicit value network.

\section{Methodology}
\label{sec:methodology}

\subsection{Balance-Oriented Inventory Allocation}
Let $\mathcal{N}=[n]$ denote the set of warehouses. For each warehouse $i\in N$, let
$I_i$ be the current inventory and $D_i$ be the forecasted daily demand. Denote $\mathcal{N}^{+}(\mathbf D)=\{i\in N: D_i>0\}$, and $\mathcal{N}^{0}(\mathbf D)=\{i\in N: D_i=0\}$ with $\mathbf D = (D_1, \ldots, D_n)^\top$.
Given a total replenishment quantity $R>0$, the decision is a nonnegative integer allocation plan $x_i\in \mathbb{Z}_{\ge 0}$ for each selected warehouse $i$, satisfying 
$\sum_{i\in\mathcal{N}} x_i = R$.
Target inventory days (TID)  is computed as
$\tau_i = \frac{I_i+x_i}{D_i}$
for each $i\in \mathcal{N}^{+}(\mathbf D)$, and 
$\tau_{\rm all}=\frac{\sum_{i\in \mathcal{N}} I_i + R}{\sum_{i\in \mathcal{N}} D_i}$
for all warehouses, respectively\footnote{Throughout this paper, we assume that at least one warehouse has strictly positive forecasted
daily demand.}.
Specifically, the warehouse $i\in\mathcal{N}^+(\mathbf D)$ is called $(\underline{\ell},\overline{\ell})$-accurately allocated if 
\[
\underline{\ell}\,\tau_{\mathrm{all}} \le \tau_i \le \overline{\ell}\,\tau_{\mathrm{all}},
\]
where $0< \underline{\ell}\le 1\le \overline{\ell}$. Furthermore, the multi-warehouse allocation accuracy\footnote{The allocation plan is constructed using forecast demand, whereas its
primary ex-post accuracy is evaluated using realized demand observed after the allocation decision.} is defined as
\[
\mathrm{Acc}_{\underline{\ell},\,\overline{\ell}}(\mathbf x;\mathbf D)
:=
\left.
\sum\limits_{\substack{
i\in\mathcal{N}^{+}(\mathbf{D}) :\\
\underline{\ell}\tau_{\mathrm{all}}
\le \tau_i
\le
\overline{\ell}\tau_{\mathrm{all}}
}}
x_i\middle/
\sum_{i\in\mathcal{N}}x_i
\right.
,\]
where $\mathbf x = (x_1, \ldots, x_n)^\top$, $\mathbf D = (D_1,\ldots,D_n)^\top$.
Similar to \citep{xu2026orla}, our goal is to achieve TID-balanced multi-warehouse inventory allocation under heterogeneous side constraints.

\subsection{Instance-Wise OR Expert Selection}

Let $\mathcal{E}=\{E_1,\ldots,E_K\}$ be the candidate OR expert library. 
Each expert corresponds to a MIP formulation with a distinct optimization priority.
For each inventory allocation instance $q$ and expert $E\in\mathcal E$, the expert first
constructs an allocation plan $\mathbf x_q^E$ using forecast demand. We define
its primary ex-post score as
\begin{equation*}
s_q(E)
=
\operatorname{Acc}
\left(\mathbf x_q^E;\mathbf D_q^{\mathrm{real}}\right),
\end{equation*}
where $\mathbf D_q^{\mathrm{real}}$ denotes the realized demand.
Larger values indicate better realized allocation quality.
It is worth noting that all candidate experts construct their allocation plans using forecast demand. Unless otherwise specified, all allocation scores, expert rankings, oracle labels, and reported evaluation metrics are subsequently computed using realized demand.
The LLM selector defines a generative 
policy $\pi_\theta(\cdot\mid q)$ that produces a structured response 
$o$, from which the selected expert is parsed:
\[
o\sim\pi_\theta(\cdot\mid q), \qquad 
\hat{E}(o)=\operatorname{Parse}(o)\in\mathcal{E}.
\]
The objective is to maximize the expected realized score of the selected 
expert:
\begin{align*}
\max_{\theta}
\mathbb{E}_{q\sim\mathcal{D},\,o\sim\pi_\theta(\cdot\mid q)}
\left(
s_q(\hat{E}(o))
\right).
\end{align*}
We instantiate the OR expert family with four complementary MIP formulations: lexicographic band (LB), scalarized band-deviation (SBD), deviation minimization (DM), and reliability-calibrated band (RCB) experts, as summarized in Table~\ref{tab:expert_library}. 
Complete formulations of the candidate OR experts are provided
in Appendix A.

\begin{table}[H]
\centering
\caption{Candidate OR expert library.}
\label{tab:expert_library}
\small
\setlength{\tabcolsep}{4pt}
\renewcommand{\arraystretch}{1.12}
\begin{tabularx}{\columnwidth}{l X}
\toprule
\multicolumn{1}{c}{Expert} & \multicolumn{1}{c}{Optimization priority} \\
\midrule

LB
& Lexicographically maximizes replenishment allocated within
the target TID band and then minimizes residual deviation. \\
\midrule

SBD
& Jointly optimizes in-band replenishment and allocation
deviation through a weighted scalarized objective. \\
\midrule

DM
& Minimizes the aggregate absolute warehouse-level TID
deviation from the system-wide TID. \\
\midrule

RCB
& Minimizes reliability-weighted TID deviation using
warehouse-level forecast variability. \\

\bottomrule
\end{tabularx}
\end{table}

\section{Solver-Guided Progressive Post-Training}
\label{sec:post-training}

\begin{figure*}[t]
\centering
\includegraphics[width=1\linewidth]{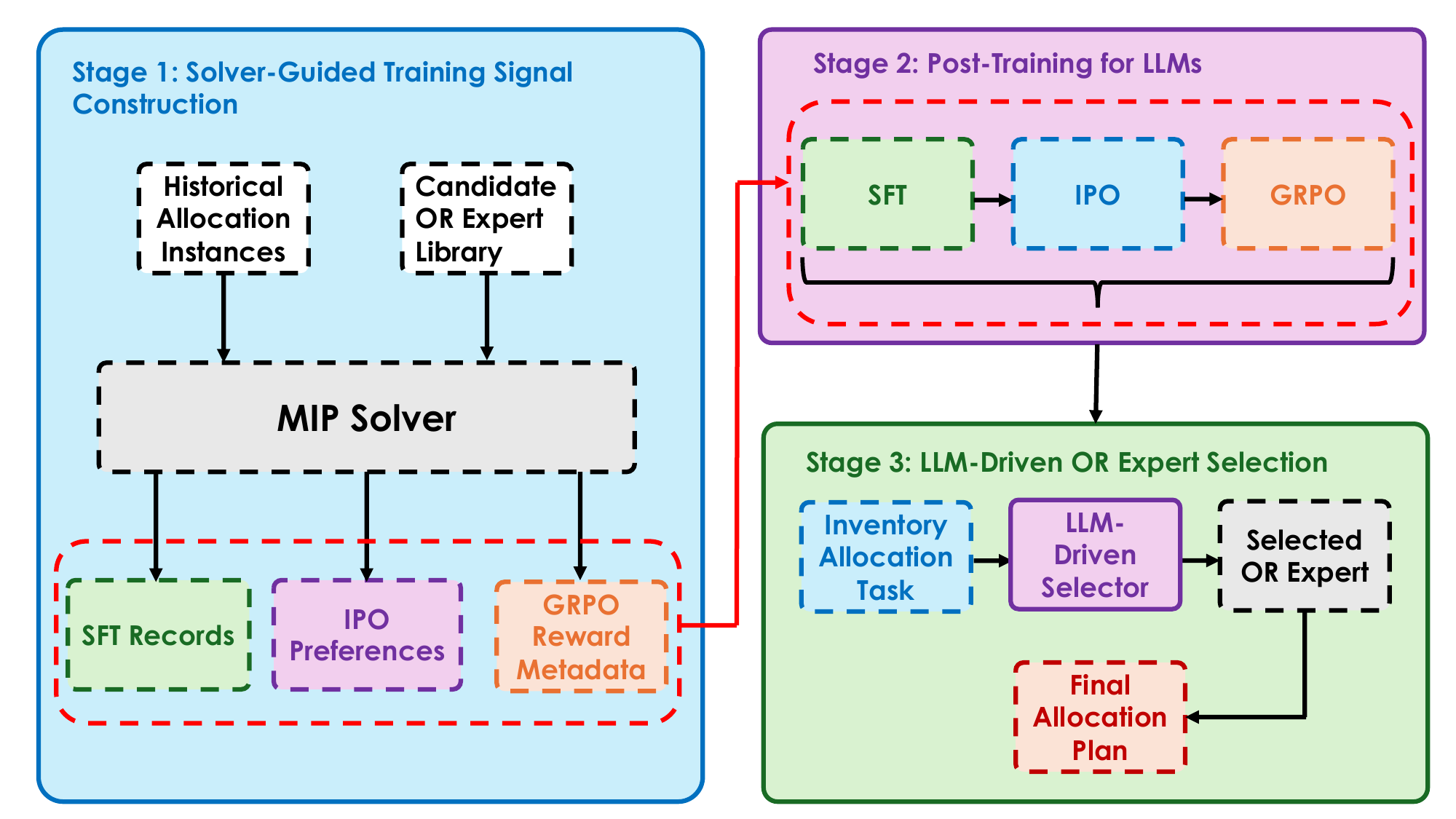}
\caption{Overview of the formulation-selection pipeline.}
\label{fig:overview_of_pipeline}
\end{figure*}

Figure~\ref{fig:overview_of_pipeline} summarizes the overall
formulation-selection pipeline.
Historical allocation instances and the candidate OR expert library
are first evaluated by a MIP solver. The resulting expert-wise scores
support three successive post-training stages. SFT first grounds the LLM in schema-consistent and balanced expert-specific formulations. IPO improves pairwise
expert discrimination using score gaps. GRPO
further optimizes the policy on its own sampled responses by querying
precomputed per-instance reward metadata.
Representative record schemas for the three post-training stages are provided in Appendix B.

\subsection{Supervised Fine-Tuning}
We first apply SFT to initialize the LLM for structured
OR-formulation selection. Each training instance is represented
as a textual instruction containing the allocation state, relevant
operational constraints, and the candidate OR expert set. The
instance information is organized into a structured representation
that enables the model to assess the compatibility between the
current allocation context and the optimization priorities encoded
by different OR experts. The output follows a unified schema
consisting of the selected OR expert and a nontrivial justification of sufficient length.
The four OR experts are sampled in a balanced proportion during SFT, encouraging the model to obtain balanced exposure to all candidate experts and learn schema-consistent selection behavior.
The resulting policy $\pi_{\mathrm{SFT}}$ is used to initialize the
subsequent IPO and GRPO stages.

\subsection{Identity Preference Optimization}
We employ Identity Preference Optimization (IPO) to enhance OR expert discrimination after SFT. 
For each instance, candidate experts are compared against the LB reference expert using solver-guided posterior quality. A preference pair is retained only when the performance gap exceeds a minimum relative score-gap threshold of 5\% and the preferred expert meets a basic quality requirement, ensuring that the chosen--rejected contrast reflects a meaningful optimization-quality difference rather than random variation. We further weight pairs by their performance margins and balance the chosen expert categories through down-sampling, preventing the IPO data from being dominated by a single expert family.

\subsection{Group Relative Policy Optimization}

\begin{figure*}[t]
\centering
\includegraphics[width=1\linewidth]{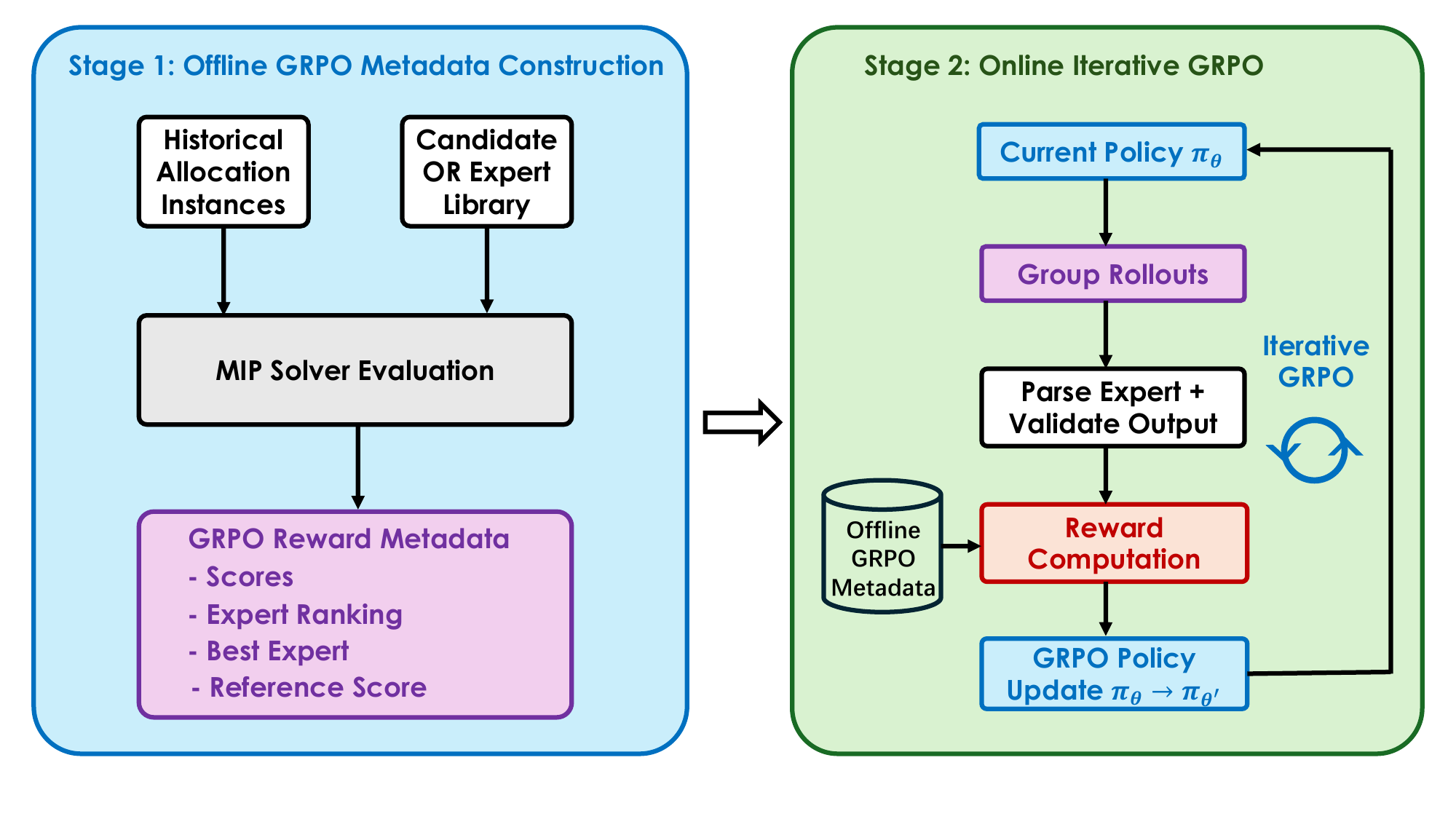}
\caption{GRPO pipeline with offline metadata construction and iterative policy optimization.}
\label{fig:overview_of_GRPO}
\end{figure*}

Building on the IPO-initialized policy $\pi_{\mathrm{IPO}}$, we
further optimize the selector using GRPO. As illustrated
in Figure~\ref{fig:overview_of_GRPO}, the procedure consists of two
stages: offline construction of per-instance reward metadata and
iterative GRPO.

\paragraph{Offline GRPO metadata construction.}
For each training instance $q$, all candidate experts are evaluated
once by the MIP solver. We store
\begin{equation*}
\mathcal{M}_q
=
\left(
\mathbf s_q,
\rho_q,
E_q^\star,
s_q^\star,
E^{\mathrm{ref}},
s_q^{\mathrm{ref}}
\right),
\end{equation*}
where $\mathbf s_q=\{s_q(E)\}_{E\in\mathcal E}$ contains the
expert-wise scores, $\rho_q$ is their descending ranking,
$s_q^\star=\max_{E\in\mathcal E}s_q(E)$,
$E_q^\star$ is the best expert, and
$s_q^{\mathrm{ref}}=s_q(E^{\mathrm{ref}})$ is the score of the
reference expert. We use LB as the reference expert because it is the
best-performing fixed formulation. The metadata are computed before
online GRPO training and cached for reward lookup.

\paragraph{Online group rollout.}
For the allocation instance $q$, the current policy samples a
group of structured responses
\begin{equation*}
\mathcal{O}_q
=
\{o_{q,g}\}_{g=1}^{G},
\qquad
o_{q,g}\sim\pi_\theta(\cdot\mid q).
\end{equation*}
Each response contains a selected expert and a nontrivial justification of sufficient length.
We parse
\begin{equation*}
E_{q,g}
=
\operatorname{Parse}(o_{q,g})
\in
\mathcal E\cup\{\bot\},
\end{equation*}
where $\bot$ denotes a malformed response or an invalid expert name.
Invalid responses receive a negative validity reward. For a valid
response, its score is retrieved from the stored
metadata:
\begin{equation*}
s_{q,g}
=
s_q(E_{q,g}).
\end{equation*}

\paragraph{Reward and policy update.}
The rollout reward combines normalized allocation quality,
ranking-aware feedback, exact-best-expert supervision,
reference-sticking regularization, and lightweight output-structure
terms. We additionally evaluate the same allocation plan using forecast
demand, yielding an auxiliary forecast-view score.
The detailed reward definitions are provided in Appendix C. Rewards and bounded token-level shaping terms are normalized
within the rollout group to construct token-level advantages.

\section{Experiments}
\label{sec:experiments}

This section evaluates the proposed formulation-selection approach on real-world instances from JD.com along two complementary dimensions: instance-wise OR expert-selection accuracy and the realized allocation quality produced by the selected expert. For each allocation instance, we solve all candidate OR formulations offline using a MIP solver and designate the expert with the highest realized allocation score as the ex-post oracle expert.

\subsection{Experimental Setup}

\paragraph{Data.}
After data filtering and balancing, the post-training datasets
comprise approximately 17,000 SFT records, 4,000 IPO
preference pairs, and 2,300 GRPO training prompts with $G$
rollouts sampled online from the current policy for
each GRPO prompt. 
Qwen3-14B\footnote{\url{https://huggingface.co/Qwen/Qwen3-14B}}
is used as the base model for all three post-training stages.
We evaluate the trained formulation
selectors on a set of 718 real-world multi-warehouse
inventory allocation instances.

\paragraph{Compared methods.}
The compared methods include the LLM selector trained with SFT+IPO, the
selector further optimized with GRPO, and the per-instance
ex-post oracle expert. For each LLM-based selector, the
generated response is first parsed to identify a candidate OR
expert. 
The selected MIP formulation is then solved using SCIP\footnote{\url{https://www.scipopt.org}} to
produce the final allocation plan.

\paragraph{Evaluation metrics.}
We report four evaluation metrics. For each instance, all four
candidate experts are ranked according to their realized
allocation accuracy. \textbf{Hit Ratio@1 (HR@1)} indicates
whether the expert selected by the LLM matches the top-ranked
expert, while \textbf{Hit Ratio@2 (HR@2)} indicates whether
the selected expert belongs to the top two experts.
\textbf{Allocation Accuracy Gain over the Incumbent Baseline (Acc. Gain)} measures
the improvement in realized allocation accuracy over the
incumbent replenishment baseline, reported in percentage points.
\textbf{Ex-post Oracle Gap (Oracle Gap)} measures the allocation-accuracy
difference between the selected expert and the per-instance ex-post best
expert, with a smaller value indicating better allocation quality.

\subsection{Main Results}
Tables~\ref{tab:expert_selection} and~\ref{tab:allocation_quality} report the main validation results in terms of expert-selection accuracy and allocation quality, respectively. 
As shown in Table~\ref{tab:expert_selection}, compared with the SFT+IPO selector, the SFT+IPO+GRPO selector achieves substantial improvements in expert-selection accuracy. Specifically, GRPO improves HR@1 from 21.45\% to 50.42\% (+28.97 pp) and HR@2 from 70.47\% to 82.31\% (+11.84 pp). 
These results indicate that GRPO not only improves exact identification of the oracle best expert, but also makes the selected expert more likely to fall within the top-two experts.

\begin{table}[t]
\centering
\caption{Expert-selection accuracy against the expert ranking on multi-warehouse inventory allocation instances. For each instance, all four candidate formulations are solved and ranked by their realized allocation accuracy. HR@1 measures whether the single expert selected by the LLM matches the top-ranked expert, while HR@2 measures whether the selected expert belongs to the top two experts in this ranking.}
\label{tab:expert_selection}
\small
\begin{tabular}{lcc}
\toprule
\textbf{Method} & 
\textbf{HR@1 (\%)} & 
\textbf{HR@2 (\%)} \\
\midrule
SFT+IPO selector 
& 21.45 
& 70.47 \\

\textbf{SFT+IPO+GRPO selector} 
& \textbf{50.42} 
& \textbf{82.31} \\
\bottomrule
\end{tabular}
\end{table}

Consistent with the improvements in expert selection, Table~\ref{tab:allocation_quality} shows that SFT+IPO+GRPO also achieves the best realized allocation quality among non-oracle methods. It obtains an allocation accuracy gain of 12.57 pp over the incumbent baseline, outperforming both SFT+IPO, which obtains 9.89 pp, and the fixed LB expert, which obtains 10.53 pp. Meanwhile, the Oracle Gap decreases from 7.53 pp under SFT+IPO to 4.85 pp after GRPO. 
These results show that GRPO not only improves exact identification of the oracle best expert, but also increases the probability that the selected expert is near-oracle, i.e., among the two best experts under the ranking.
\begin{table}[t]
\centering
\caption{Allocation-quality comparison on multi-warehouse inventory
allocation instances. Acc. Gain denotes the allocation accuracy
improvement over the incumbent baseline, while Oracle Gap denotes
the difference from the per-instance ex-post oracle expert. Both are
reported in percentage points.}

\label{tab:allocation_quality}
\small
\begin{tabular}{lcc}
\toprule
\textbf{Method} & 
\textbf{Acc. Gain (pp)} & 
\textbf{Oracle Gap (pp)} \\
\midrule
LB
& 10.53 
& 6.89 \\

SFT+IPO 
& 9.89 
& 7.53 \\

\textbf{SFT+IPO+GRPO} 
& \textbf{12.57} 
& \textbf{4.85} \\

Oracle best expert 
& 17.42 
& 0.00 \\
\bottomrule
\end{tabular}
\end{table}
Overall, these results indicate that GRPO strengthens instance-level expert selection and improves realized allocation performance.

\subsection{Oracle-Conditioned Selection Analysis}

To further analyze where the GRPO gains come from, we group validation instances according to their ex-post oracle best expert and compute HR@1 and HR@2 within each oracle bucket.

\begin{figure*}[t]
\centering

\begin{subfigure}[t]{0.48\textwidth}
    \centering
    \includegraphics[width=\linewidth]{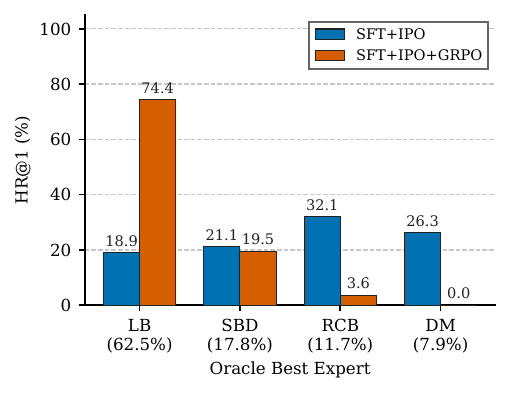}
    \caption{HR@1 by oracle-best bucket.}
    \label{fig:hit1_by_bucket}
\end{subfigure}
\hfill
\begin{subfigure}[t]{0.48\textwidth}
    \centering
    \includegraphics[width=\linewidth]{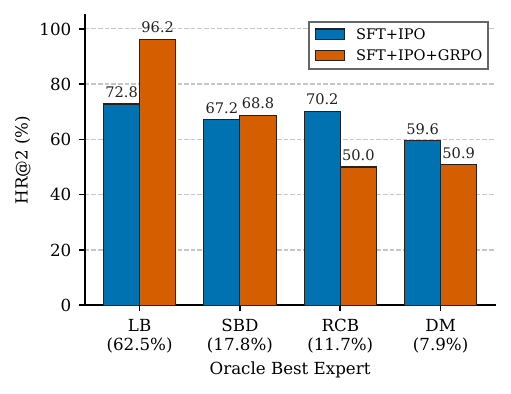}
    \caption{HR@2 by oracle-best bucket.}
    \label{fig:hit2_by_bucket}
\end{subfigure}

\caption{Oracle-conditioned expert-selection accuracy.
Instances are grouped by their solver-derived oracle best expert.
HR@1 measures whether the LLM-selected expert matches the best expert, while HR@2 measures whether the selected expert belongs to the top two experts.}
\label{fig:grpo_reward}
\end{figure*}

Figure \ref{fig:grpo_reward} shows that the overall HR@1 improvement of GRPO is mainly driven by the LB bucket, which accounts for the dominant portion of the validation set. In this bucket, SFT+IPO+GRPO improves HR@1 from 18.9\% to 74.4\% and HR@2 from 72.8\% to 96.2\%. This indicates that GRPO effectively corrects expert selection in the dominant oracle regime.

For the SBD bucket, the effect of GRPO is more moderate. HR@1 slightly decreases from 21.1\% to 19.5\%, while HR@2 increases from 67.2\% to 68.8\%. 
This suggests that, for this regime, GRPO does not always identify the oracle best expert exactly, but its selected expert more often remains within the top-two experts.
By contrast, GRPO shows limited effectiveness on the minority buckets RCB and DM. 
For RCB, HR@1 decreases from 32.1\% to 3.6\%, while HR@2 decreases from 70.2\% to 50.0\%. For DM, HR@1 decreases from 26.3\% to 0.0\%, and HR@2 decreases from 59.6\% to 50.9\%.
This pattern suggests that GRPO primarily improves overall utility on the observed instance distribution, but does not guarantee balanced expert recognition across oracle regimes. 
In other words, GRPO improves overall selection utility and top-ranked oracle matching in the dominant regime, while still exhibiting under-selection for regimes with fewer samples or sparser structural signals.

This diagnostic result clarifies both the benefit mechanism of the
solver-guided training procedure and an important direction for
future improvement. The current GRPO training objective tends to optimize overall allocation quality, so its gains are concentrated in frequent and high-impact oracle regimes. Future work may incorporate class-balanced rewards or minority-regime oversampling to improve expert selection for minority regimes.

\section{Conclusions}
\label{sec:conclusions}

Instead of treating formulation design as a fixed modeling choice,
the proposed framework selects a MIP expert according
to the structural characteristics of each allocation instance.
Posterior solver evidence is converted into progressive supervision
for the LLM-based formulation selector. SFT establishes
schema-consistent generation. IPO introduces pairwise discrimination
from realized quality gaps. GRPO further aligns the selector with
allocation utility through within-instance relative rewards.

Experimental results show that formulation selection is not only a
prediction problem over expert labels, but also a utility-driven
decision problem whose value should be assessed after optimization.
The resulting SFT+IPO+GRPO selector improves allocation quality over
both fixed-formulation and preference-trained baselines. Meanwhile,
the oracle-conditioned analysis reveals that most gains are
concentrated in dominant expert regimes.

These findings suggest several directions for future work. First,
class-balanced rewards, minority-regime oversampling, or cost-sensitive
objectives may improve recognition of sparse regimes. Second, the
expert library can be expanded to cover richer operational scenarios
and uncertainty structures. Third, online solver feedback and
deployment-aware reward design may further close the gap between
offline validation and production allocation systems.

\bibliography{aaai2027}

\begin{thebibliography}{16}
\providecommand{\natexlab}[1]{#1}

\bibitem[{Ali et~al.(2025)Ali, Ashraf, Hegazy, Salem, Mokhtar, Gaber, and
  Alrefaie}]{ali2025pair}
Ali, S.; Ashraf, M.; Hegazy, S.; Salem, F.; Mokhtar, H.; Gaber, M.~M.; and
  Alrefaie, M.~T. 2025.
\newblock PAIR: A Novel Large Language Model-Guided Selection Strategy for
  Evolutionary Algorithms.
\newblock \emph{arXiv preprint arXiv:2503.03239}.

\bibitem[{Astorga et~al.(2025)Astorga, Liu, Xiao, and Van
  Der~Schaar}]{Astorga2024}
Astorga, N.; Liu, T.; Xiao, Y.; and Van Der~Schaar, M. 2025.
\newblock Autoformulation of Mathematical Optimization Models Using {LLM}s.
\newblock In Singh, A.; Fazel, M.; Hsu, D.; Lacoste-Julien, S.; Berkenkamp, F.;
  Maharaj, T.; Wagstaff, K.; and Zhu, J., eds., \emph{Proceedings of the 42nd
  International Conference on Machine Learning}, volume 267 of
  \emph{Proceedings of Machine Learning Research}, 1864--1886. PMLR.

\bibitem[{{DeepSeek-AI}(2025)}]{deepseekai2025r1}
{DeepSeek-AI}. 2025.
\newblock DeepSeek-R1: Incentivizing Reasoning Capability in LLMs via
  Reinforcement Learning.
\newblock \emph{arXiv preprint arXiv:2501.12948}.

\bibitem[{Du et~al.(2024)Du, Liu, Lin, Niyato, Kang, Xiong, and
  Kim}]{du2024moe_network_opt}
Du, H.; Liu, G.; Lin, Y.; Niyato, D.; Kang, J.; Xiong, Z.; and Kim, D.~I. 2024.
\newblock Mixture of Experts for Network Optimization: A Large Language
  Model-enabled Approach.
\newblock \emph{arXiv preprint arXiv:2402.09756}.

\bibitem[{Gao et~al.(2025)Gao, Shang, Xue, and Qian}]{gao2025neural}
Gao, C.; Shang, H.; Xue, K.; and Qian, C. 2025.
\newblock Neural Solver Selection for Combinatorial Optimization.
\newblock In \emph{Proceedings of the 42nd International Conference on Machine
  Learning}, volume 267 of \emph{Proceedings of Machine Learning Research},
  18528--18549. PMLR.

\bibitem[{Gheshlaghi~Azar et~al.(2024)Gheshlaghi~Azar, Daniel~Guo, Piot, Munos,
  Rowland, Valko, and Calandriello}]{GheshlaghiAzar2024}
Gheshlaghi~Azar, M.; Daniel~Guo, Z.; Piot, B.; Munos, R.; Rowland, M.; Valko,
  M.; and Calandriello, D. 2024.
\newblock A General Theoretical Paradigm to Understand Learning from Human
  Preferences.
\newblock In Dasgupta, S.; Mandt, S.; and Li, Y., eds., \emph{Proceedings of
  The 27th International Conference on Artificial Intelligence and Statistics},
  volume 238 of \emph{Proceedings of Machine Learning Research}, 4447--4455.
  PMLR.

\bibitem[{Huang et~al.(2025)Huang, Tang, Hu, Jiang, Zheng, Ge, Wang, and
  Wang}]{Huang2025}
Huang, C.; Tang, Z.; Hu, S.; Jiang, R.; Zheng, X.; Ge, D.; Wang, B.; and Wang,
  Z. 2025.
\newblock {ORLM}: A Customizable Framework in Training Large Models for
  Automated Optimization Modeling.
\newblock \emph{Operations Research}, 73(6): 2986--3009.

\bibitem[{Jiang et~al.(2025)Jiang, Shu, Qian, Lu, Zhou, Zhou, and
  Yu}]{Jiang2024}
Jiang, C.; Shu, X.; Qian, H.; Lu, X.; Zhou, J.; Zhou, A.; and Yu, Y. 2025.
\newblock {LLMOPT}: Learning to Define and Solve General Optimization Problems
  from Scratch.
\newblock In \emph{The Thirteenth International Conference on Learning
  Representations (ICLR 2025)}.

\bibitem[{Ouyang et~al.(2022)Ouyang, Wu, Jiang, Almeida, Wainwright, Mishkin,
  Zhang, Agarwal, Slama, Ray, Schulman, Hilton, Kelton, Miller, Simens, Askell,
  Welinder, Christiano, Leike, and Lowe}]{ouyang2022training}
Ouyang, L.; Wu, J.; Jiang, X.; Almeida, D.; Wainwright, C.~L.; Mishkin, P.;
  Zhang, C.; Agarwal, S.; Slama, K.; Ray, A.; Schulman, J.; Hilton, J.; Kelton,
  F.; Miller, L.; Simens, M.; Askell, A.; Welinder, P.; Christiano, P.; Leike,
  J.; and Lowe, R. 2022.
\newblock Training language models to follow instructions with human feedback.
\newblock In Koyejo, S.; Mohamed, S.; Agarwal, A.; Belgrave, D.; Cho, K.; and
  Oh, A., eds., \emph{Advances in Neural Information Processing Systems},
  volume~35, 27730--27744. Curran Associates, Inc.
\newblock ISBN 9781713871088.

\bibitem[{Schulman et~al.(2017)Schulman, Wolski, Dhariwal, Radford, and
  Klimov}]{schulman2017ppo}
Schulman, J.; Wolski, F.; Dhariwal, P.; Radford, A.; and Klimov, O. 2017.
\newblock Proximal Policy Optimization Algorithms.
\newblock \emph{arXiv preprint arXiv:1707.06347}.

\bibitem[{Shao et~al.(2024)Shao, Wang, Zhu, Xu, Song, Bi, Zhang, Zhang, Li, Wu,
  and Guo}]{shao2024deepseekmath}
Shao, Z.; Wang, P.; Zhu, Q.; Xu, R.; Song, J.; Bi, X.; Zhang, H.; Zhang, M.;
  Li, Y.~K.; Wu, Y.; and Guo, D. 2024.
\newblock DeepSeekMath: Pushing the Limits of Mathematical Reasoning in Open
  Language Models.
\newblock \emph{arXiv preprint arXiv:2402.03300}.

\bibitem[{Wu et~al.(2024)Wu, Zhong, Wu, Jiang, and
  Tan}]{wu2024llm_algorithm_selection}
Wu, X.; Zhong, Y.; Wu, J.; Jiang, B.; and Tan, K.~C. 2024.
\newblock Large Language Model-Enhanced Algorithm Selection: Towards
  Comprehensive Algorithm Representation.
\newblock In \emph{Proceedings of the Thirty-Third International Joint
  Conference on Artificial Intelligence (IJCAI-24)}, 5235--5244.

\bibitem[{Xu et~al.(2026)Xu, Ma, Dong, Qi, Zhang, Geng, and Zhang}]{xu2026orla}
Xu, J.; Ma, Y.; Dong, J.; Qi, Y.; Zhang, J.; Geng, D.; and Zhang, A. 2026.
\newblock Solver-Verified Formulation Generation and Selection for
  Multi-Warehouse Inventory Allocation Using Large Language Models.
\newblock \emph{arXiv preprint arXiv:2606.29366}.

\bibitem[{Zhang, Feng, and You(2025)}]{zhang2025routerr1}
Zhang, H.; Feng, T.; and You, J. 2025.
\newblock Router-{R1}: Teaching {LLMs} Multi-Round Routing and Aggregation via
  Reinforcement Learning.
\newblock In Belgrave, D.; Zhang, C.; Lin, H.; Pascanu, R.; Koniusz, P.;
  Ghassemi, M.; and Chen, N., eds., \emph{Advances in Neural Information
  Processing Systems}, volume~38, 141233--141265. Curran Associates, Inc.

\bibitem[{Zhang and Yi(2025)}]{zhang2025laos}
Zhang, Y.; and Yi, G. 2025.
\newblock LAOS: Large Language Model-Driven Adaptive Operator Selection for
  Evolutionary Algorithms.
\newblock In \emph{Proceedings of the Genetic and Evolutionary Computation
  Conference (GECCO '25)}, 517--526. Association for Computing Machinery.

\bibitem[{Zhou et~al.(2026)Zhou, Tan, Zhang, Fan, Lin, Kang, Song, Li, Huang,
  Yu, Fan, Chen, Xu, Liu, Qin, Torr, Zhang, and Yin}]{zhou2026select}
Zhou, H.; Tan, Z.; Zhang, Z.; Fan, Y.; Lin, Y.; Kang, L.; Song, X.; Li, R.;
  Huang, S.; Yu, A.; Fan, Y.; Chen, Y.; Xu, K.; Liu, X.; Qin, Y.; Torr, P.;
  Zhang, C.; and Yin, Z. 2026.
\newblock Select-then-Solve: Paradigm Routing as Inference-Time Optimization
  for LLM Agents.
\newblock \emph{arXiv preprint arXiv:2604.06753}.

\end{thebibliography}

\appendix

\section{A~~~Formulations of the Candidate OR Experts}
\label{app:MIP_expert}

This section presents the complete formulations of the four
candidate OR experts. They share the same replenishment-
conservation structure but encode different allocation priorities,
forming the expert library used by the LLM selector.

\subsection{A.1 Lexicographic Band Expert}
For each $i\in \mathcal{N}^{+}(\mathbf D)$, we define the target allocation
\[
\hat{x}_i=\max\{0,\; \tau_{all} D_i - I_i\},
\]
and the admissible allocation band
\[
l_i=\max\{0, \left\lceil \underline{\ell} \tau_{all} D_i - I_i \right\rceil\},~~~
u_i=\max\{0, \left\lfloor \overline{\ell} \tau_{all} D_i - I_i \right\rfloor\}.
\]
Let $\mathcal{F}=\{i\in\mathcal{N}^{+}(\mathbf D): u_i\ge l_i\}$.

For each warehouse $i$, auxiliary decision variable $z_i\in\{0,1\}$ 
indicates whether $i$ is counted as lying in the target band, $y_i$ denotes the replenishment volume that is counted as being allocated within the target band. Let
\begin{align*}
\Omega=
\left\{
(\mathbf{x},\mathbf{y},\mathbf{z}) \left|
\begin{array}{ll}
\sum_{i\in \mathcal{N}} x_i = R, \\[1mm]
l_i z_i \le x_i \le u_i z_i + R(1-z_i),  \forall i\in \mathcal{F}, \\[1mm]
y_i = x_i z_i, \forall i \in \mathcal{F}, \\[1mm]
z_i = 0, \forall i\in (\mathcal{N}^{+}(\mathbf D)\setminus \mathcal{F})\cup \mathcal{N}^{0}(\mathbf D), \\
y_i = 0, \forall i\in (\mathcal{N}^{+}(\mathbf D)\setminus \mathcal{F})\cup \mathcal{N}^{0}(\mathbf D),\\
x_i\in \mathbb{Z}_{\ge 0},\; y_i\ge 0,\; z_i\in\{0,1\},\;   \forall i\in \mathcal{N} 
\end{array}\right.
\right\}.
\end{align*}
With the above definitions, lexicographic band (LB) expert solves a lexicographic OR problem over $\Omega$. 
It first maximizes the on-target allocation volume and then minimizes allocation deviation among Stage-1 optimal solutions.

\noindent \emph{Stage 1: Maximizing on-target replenishment volume.}
\begin{align*}
f_1^* =  \max_{(\mathbf x,\mathbf y, \mathbf z)\in\Omega} \quad & \sum_{i\in \mathcal{N}} y_i. \label{eq:stage1_obj}
\end{align*}

\noindent \emph{Stage 2: Deviation minimization under stage-1 optimality.}
Let
\begin{align*}
\Omega^* = \left\{(\mathbf x,\mathbf y, \mathbf z)\in\Omega\left|\sum_{i\in\mathcal{N}}y_i = f_1^*\right.\right\},
\end{align*}
then we solve
\begin{align*}
\min_{(\mathbf x, \mathbf y, \mathbf z)\in\Omega^*} \quad & \sum_{i\in \mathcal{N}^{+}(\mathbf D)} \vert x_i - \hat x_i\vert \; + \; \epsilon \sum_{i\in \mathcal{N}^{0}(\mathbf D)} x_i.
\end{align*}

\paragraph{Linearized reformulation.}
Note that $\sum_{i\in\mathcal{N}}x_i=R$ and $x_i\geq 0$
imply $0\leq x_i\leq R$, the bilinear equality
$y_i=x_i z_i$ can be exactly replaced, for each $i\in \mathcal{F}$, by
\begin{align*}
0 \leq y_i \leq x_i,\quad
y_i \leq Rz_i,\quad
y_i \geq x_i-R(1-z_i).
\end{align*}
To linearize the absolute-deviation terms in Stage~2, we
introduce continuous auxiliary variables $d_i\geq 0$ satisfying
\begin{equation*}
d_i\geq x_i-\hat{x}_i,\quad
d_i\geq \hat{x}_i-x_i,
\quad i\in\mathcal{N}^{+}(\mathbf D).
\end{equation*}
The Stage~2 objective is consequently written as
\begin{align*}
\sum_{i\in\mathcal{N}^{+}(\mathbf D)}d_i+
\epsilon\sum_{i\in\mathcal{N}^{0}(\mathbf D)}x_i.
\end{align*}
Hence, the LB expert is solved exactly as two sequential
mixed-integer linear programming (MILP) problems implementing the lexicographic objective.

\subsection{A.2 Scalarized Band-Deviation Expert}

The scalarized band-deviation (SBD) formulation is shown as below, which can be equivalently reformulated as a MILP problem \citep{xu2026orla}.
\begin{align*}
\max_{\mathbf x,\, \mathbf y\, ,\mathbf z} \quad
& \lambda_1 \sum_{i\in\mathcal{N}} y_i
- \lambda_2 \sum_{i\in\mathcal{N}}
\left| I_i + x_i - T_i \right| \\
\text{s.t.} \quad
& \sum_{i\in\mathcal{N}} x_i = R, \\
& I_i + x_i
\ge \underline{\ell} \,T_i - M(1-z_i),
\quad \forall i\in\mathcal{N}, \\
& I_i + x_i
\le \overline{\ell} \,T_i + M(1-z_i),
\quad \forall i\in\mathcal{N}, \\
& y_i = x_i z_i,\quad \forall i\in\mathcal{N}\\
& y_i \ge 0,\quad
x_i \in \mathbb{Z}_{\ge 0},\quad
z_i \in \{0,1\},
\quad \forall i\in\mathcal{N},
\end{align*}
where $T_i = \tau_{\rm all}D_i, \forall i\in\mathcal{N}$, $\lambda_1, \lambda_2>0$ control the trade-off
between in-band replenishment and allocation deviation, and
$M>0$ is a sufficiently large constant.

\subsection{A.3 Deviation Minimization Expert}
This deviation minimization (DM) formulation is introduced as below, which can be also equivalently reformulated as a MILP problem \citep{xu2026orla}.
\begin{align*}
\min_{\mathbf x}\quad
& \sum_{i\in\mathcal{N}^+(\mathbf D)}
\left|
\frac{I_i+x_i}{D_i} - \tau_{\rm all}
\right| \\
\text{s.t.}\quad
& \sum_{i\in\mathcal{N}} x_i = R, \\
& x_i \in \mathbb{Z}_{\ge 0},\quad \forall i\in\mathcal{N}.
\end{align*}

\subsection{A.4 Reliability-Calibrated Band Expert} 
Reliability-calibrated band (RCB) expert accounts for heterogeneous forecast reliability by weighting the
TID deviation of each warehouse according to its forecast variability. We define a warehouse-specific uncertainty-adjusted TID band $[L_i,U_i]$ by
\begin{align*}
U_i =
\frac{\sum_{j\in\mathcal{N}}I_j+R}
{\sum_{j\in\mathcal{N}}D_j},
\qquad
L_i =
\frac{\sum_{j\in\mathcal{N}}I_j+R}
{\sum_{j\in\mathcal{N}}D_j+m\sigma_i}
\end{align*}
for each $i\in \mathcal{N}^+(\mathbf D)$, where \(m\) controls the uncertainty inflation. The model is
{\footnotesize\begin{align*}
\min_{\mathbf x}\quad
& \sum_{i\in\mathcal{N}^+(\mathbf D)}
\frac{1}{CV_i + \delta_{\rm CV}}\left(\left|
\frac{I_i+x_i}{D_i} - L_i
\right| + \left|
\frac{I_i+x_i}{D_i} - U_i
\right|\right) \\
\text{s.t.}\quad
& \sum_{i\in\mathcal{N}} x_i = R, \\
& x_i \in \mathbb{Z}_{\ge 0},\quad i\in\mathcal{N},
\end{align*}}

\noindent where $CV_i$ denotes the coefficient of variation of the
forecast demand sequence, and $\delta_{\rm CV}>0$ is a small constant. This
weighting scheme imposes tighter balancing on warehouses with reliable
forecasts and relaxes the pressure on warehouses with volatile demand.

\paragraph{Linearized reformulation.}
For each $i\in\mathcal{N}^{+}(\mathbf D)$, introduce nonnegative
continuous variables $d_i^{L}$ and $d_i^{U}$ satisfying
\begin{equation*}
\begin{aligned}
d_i^{L} &\geq I_i+x_i-D_iL_i, \quad
d_i^{L} \geq D_iL_i-I_i-x_i,\\
d_i^{U} &\geq I_i+x_i-D_iU_i, \quad
d_i^{U} \geq D_iU_i-I_i-x_i.
\end{aligned}
\end{equation*}
The objective function is then equivalently written as
\begin{equation*}
\sum_{i\in\mathcal{N}^{+}(\mathbf D)}
\frac{d_i^{L}+d_i^{U}}
{(CV_i+\delta_{\mathrm{CV}})D_i}.
\end{equation*}

\section{B~~~Post-Training Schemas}
\label{app:training_schemas}

This appendix presents representative record schemas used in
the three post-training stages. 
Figure~\ref{fig:sft_framework}
illustrates an SFT record. 
Figure~\ref{fig:ipo_framework} shows
an IPO preference record containing chosen and rejected responses
constructed from score differences.
Figure~\ref{fig:grpo_framework} presents the GRPO prompt together with its associated reward metadata.

\begin{figure}[!t]
\centering
\begin{minipage}[t]{0.45\textwidth}
\begin{tcblisting}{
  instructionbox,
  title=TASK PROMPT,
  width=\linewidth,
  listing options={
    style=recordjson,
    numbers=none,
    breakautoindent=false
  }
}
{
  "instruction": "Act as a supply-chain expert specialized in formulating operations research (OR) models under the multi-warehouse  inventory allocation setting. Given the inputs, features, and candidate OR experts, assess SKU-level characteristics such as inventory imbalance and demand concentration, and identify the OR expert that best matches the current SKU instance."
}
\end{tcblisting}

\vspace{5mm}

\begin{tcblisting}{
  inputbox,
  title=INSTANCE INPUT,
  width=\linewidth
}
{
  "SKU ID": "SKU_ID_1",
  "total replenishment": 10,
  "warehouse information": [
    {
      "name": "W1",
      "forecast demand": 2,
      "current inventory": 20
    },
    {
      "name": "W2",
      "forecast demand": 5,
      "current inventory": 50
    },
    {
      "name": "W3",
      "forecast demand": 3,
      "current inventory": 35
    }
  ],
  "features":
  {
    "demand concentration": 0.2,
    "inventory imbalance": 3.5,
    "inventory skewness": 0,
    "additional features": "..."
  },
  "candidate OR experts": ["LB", "SBD", "DM", "RCB"]
}
\end{tcblisting}

\vspace{5mm}

\begin{tcblisting}{
  outputbox,
  title=TARGET RESPONSE,
  width=\linewidth
}
{
  "OR expert": "DM",
  "reason": "DM is selected because this instance is better addressed by minimizing warehouse-level deviations from the system-wide TID without explicitly prioritizing target-band membership."
}
\end{tcblisting}

\end{minipage}
\caption{Representative schema of an SFT training record.}
\label{fig:sft_framework}
\end{figure}

%%%%%%%%%%%%%%%%%%%%%%%%%%%%%%%%%%%%%%%%%%%%

\begin{figure}[!t]
\centering
\begin{minipage}[t]{0.45\textwidth}
\begin{tcblisting}{
  instructionbox,
  title=TASK PROMPT,
  width=\linewidth,
  listing options={
    style=recordjson,
    numbers=none,
    breakautoindent=false
  }
}
{
 "instruction": "Act as a supply-chain expert specialized in formulating operations research (OR) models under the multi-warehouse  inventory allocation setting. Given the inputs, features, and candidate OR experts, assess SKU-level characteristics such as inventory imbalance and demand concentration, and identify the OR expert that best matches the current SKU instance."
}
\end{tcblisting}

\vspace{7mm}

\begin{tcblisting}{
  inputbox,
  title=INSTANCE INPUT,
  width=\linewidth
}
{
  "SKU ID": "SKU_ID_2",
  "total replenishment": 25,
  "warehouse information": [
    {
      "name": "W1",
      "forecast demand": 2,
      "current inventory": 30
    },
    {
      "name": "W2",
      "forecast demand": 1,
      "current inventory": 13
    },
    {
      "name": "W3",
      "forecast demand": 3,
      "current inventory": 35
    }
  ],
  "features":
  {
    "demand concentration": 0.5,
    "inventory imbalance": 2.69,
    "inventory skewness": -0.56,
    "additional features": "..."
  },
  "candidate OR experts": ["LB", "SBD", "DM", "RCB"]
}
\end{tcblisting}

\vspace{7mm}

\begin{tcblisting}{
  chosenbox,
  title=CHOSEN RESPONSE,
  width=\linewidth
}
{
  "OR expert": "SBD",
  "reason": "SBD jointly balances in-band replenishment volume and residual allocation deviation through a weighted scalarized objective."
}
\end{tcblisting}

\vspace{7mm}

\begin{tcblisting}{
  rejectedbox,
  title=REJECTED RESPONSE,
  width=\linewidth
}
{
  "OR expert": "LB",
  "reason": "LB prioritizes in-band replenishment through a lexicographic objective before minimizing residual allocation deviation."
}
\end{tcblisting}
\end{minipage}
\caption{Representative schema of an IPO preference record.}
\label{fig:ipo_framework}
\end{figure}

%%%%%%%%%%%%%%%%%%%%%%%%%%%%%%%%%%%%%%%

\begin{figure}[!t]
\centering
\begin{minipage}[t]{0.45\textwidth}
\begin{tcblisting}{
  promptbox,
  title=TASK PROMPT,
  width=\linewidth
}
{
  "instruction": "Act as a supply-chain expert specialized in formulating operations research (OR) models under the multi-warehouse  inventory allocation setting. Given the inputs, features, and candidate OR experts, assess SKU-level characteristics such as inventory imbalance and demand concentration, and identify the OR expert that best matches the current SKU instance.",
  "allocation instance": {
    "SKU ID": "SKU_ID_3",
    "total replenishment": 10,
    "warehouse information": [
      {
        "name": "W1",
        "forecast demand": 6,
        "current inventory": 10
      },
      {
        "name": "W2",
        "forecast demand": 12,
        "current inventory": 40
      },
      {
        "name": "W3",
        "forecast demand": 10,
        "current inventory": 30
      }
    ]
  },
  "features": {
    "demand concentration": 0.3,
    "TID gap": 1.5,
    "inventory imbalance": 4,
    "inventory skewness": -0.38,
    "additional features": "..."
  },
  "candidate OR experts": ["LB", "SBD", "DM", "RCB"]
}
\end{tcblisting}

\vspace{3mm}

\begin{tcblisting}{
  metadatabox,
  title=REWARD METADATA,
  width=\linewidth
}
{
  "best expert": "SBD",
  "realized-demand expert scores":
  {
    "SBD": 0.9,
    "LB": 0.6,
    "RCB": 0.4,
    "DM": 0.3
  },
  "forecast-demand expert scores":
  {
    "SBD": 0.95,
    "LB": 0.7,
    "RCB": 0.5,
    "DM": 0.4
  },
  "expert ranking":
  [
   "SBD", "LB", "RCB", "DM"
  ],
  "reference expert": "LB",
  "reference realized-demand score": 0.6,
  "reference forecast-demand score": 0.7,
  "best realized-demand score": 0.9
}

\end{tcblisting}
\end{minipage}
\caption{Representative schema of a GRPO training record with reward metadata.}
\label{fig:grpo_framework}
\end{figure}

\section{C~~~GRPO Reward Construction}
\label{app:grpo_reward}

This appendix provides the detailed reward construction used in the
GRPO stage.

\subsection{C.1 Validity Gates} For each rollout, the response is first parsed into a structured formulation-selection output. A reward is assigned only when the selected formulation is valid and its score is available. Otherwise, the rollout receives a negative validity reward. This validity-aware design ensures that GRPO optimizes formulation selection under executable and score-grounded outputs rather than malformed generations.

\subsection{C.2 Main Reward}
For a valid
response $o_{q,g}$, the main reward is
\[
r_{q,g}^{\rm main}
:=
r_{q,g}^{\rm alloc} + r_{q,g}^{\rm rank} + r_{q,g}^{\rm target} + r_{q,g}^{\rm reason} + r_{q,g}^{\rm fore}
-
p_{q,g}^{\rm stick}.
\]

\paragraph{Allocation quality reward.}
To make rewards comparable across instances, we normalize the improvement
over the  reference expert by the attainable headroom:
\begin{align*}
\tilde r_{q,g}^{\rm alloc}
=
\rm{clip}_{[-\kappa_{\rm alloc},\kappa_{\rm alloc}]}
\left(
\frac{\Delta_{q,g}}{\eta_q}
\right),
\end{align*}
where
\[
\Delta_{q,g}:=s_{q,g}-s_q^{\rm ref},~~~
\eta_q:=\max(s_q^\star-s_q^{\rm ref}, \epsilon_\eta)
\]
with $\epsilon_{\eta}>0$, $\kappa_{\rm alloc}>0$. 

We further apply an instance-adaptive anchor to discour-
age collapse to the reference formulation when the attainable
headroom is large:
\[
r_{q,g}^{\rm alloc} = \tilde r_{q,g}^{\rm alloc} - d_q,
\]
where
\[
d_q=
\begin{cases}
d_q^{\rm high}, & \eta_q\ge \bar \eta,\\
d_q^{\rm low}, & \eta_q<  \bar \eta,
\end{cases} ~~~~d_q^{\rm high} > d_q^{\rm low}>0, \bar \eta >0. 
\]

\paragraph{Relative-performance reward.}
The relative-performance reward provides tiered coarse ranking supervision:
\[
r^{\mathrm{rank}}_{q,g}
=
\begin{cases}
\beta_{\mathrm{tie}}, & s_q^\star - s_{q,g} \le \varepsilon_{\mathrm{tie}},\\
\beta_{\mathrm{upper}}, & \Delta_{q,g} \ge \tau_{\Delta}^{\mathrm{upper}},\\
\beta_{\mathrm{lower}}, & \Delta_{q,g} \ge \tau_{\Delta}^{\mathrm{low}},\\
\beta_{\mathrm{penalty}}, & \text{otherwise},
\end{cases}
\]
where $\beta_{\mathrm{tie}}>\beta_{\mathrm{upper}}>\beta_{\mathrm{lower}}>0>\beta_{\mathrm{penalty}}$, $\varepsilon_{\mathrm{tie}}>0$, $\tau_{\Delta}^{\mathrm{upper}}>0$, $\tau_{\Delta}^{\mathrm{low}}<0$.

\paragraph{Target reward.}
The target reward encourages exact identification of the best expert
\[
r_{q,g}^{\rm target}
=
\gamma^{\rm target}\cdot \mathbf{1}\left\{E_{q,g}=E_q^\star\right\},
\]
where $\gamma^{\rm target}>0$.

\paragraph{Reference-sticking penalty.}
The reference-sticking penalty discourages the policy from choosing the reference expert when the attainable improvement over it is sufficiently large:
\[
p_{q,g}^{\rm stick}
=
\gamma^{\rm stick}\cdot
\mathbf{1}
\!\left\{
E_{q,g}=E^{\rm ref}~
{\rm and}~
\eta_q\ge \hat \eta
\right\},
\]
where $\gamma^{\rm stick}>0$, $\hat \eta>0$.

\paragraph{Reasoning-length reward.}
We add a small bonus to encourage the model to provide a valid concise
justification:
\[
r^{\mathrm{reason}}_{q,g}
=
\lambda_{\mathrm{reason}}
\cdot
\mathbf{1}
\left\{
|\texttt{reason}_{q,g}| \ge L_{\min}
\right\},
\]
where $\lambda_{\mathrm{reason}}>0$ is the bonus weight and $L_{\min}>0$
is the minimum reasoning-length threshold.

\paragraph{Forecast-view auxiliary reward.}
Following the convention in the main text, the primary expert
scores and rankings are evaluated ex-post using realized demand.
As a lightweight auxiliary signal, we additionally evaluate the same
allocation plan under forecast demand.

Let
\begin{equation*}
s_{q,g}^{\mathrm{fore}}
:=
\mathrm{Acc}^{\mathrm{fore}}_q
\left(x_q^{E_{q,g}}\right),
s_q^{\mathrm{fore,ref}}
:=
\mathrm{Acc}^{\mathrm{fore}}_q
\left(x_q^{E^{\mathrm{ref}}}\right)
\end{equation*}
denote the allocation and reference accuracy evaluated using forecast
demand. We define
\[
r^{\mathrm{fore}}_{q,g}
=
\lambda_{\mathrm{fore}}
\cdot
\operatorname{clip}_{[-\kappa_{\mathrm{fore}},\kappa_{\mathrm{fore}}]}
\left(
\frac{s^{\mathrm{fore}}_{q,g}-s^{\mathrm{fore,ref}}_q}
{\tau_{\mathrm{fore}}}
\right),
\]
where $\tau_{\mathrm{fore}}>0$ normalizes the
forecast-side score difference, $\kappa_{\mathrm{fore}}>0$ bounds the auxiliary
reward range, and $\lambda_{\mathrm{fore}}>0$ controls its weight.

\subsection{C.3 Token-Level Advantage Construction}
To align rollout-level supervision with token-level policy optimization, the scalar reward of each rollout is first broadcast to all response tokens. We then apply a lightweight shaping strategy only to selection-critical fields. Tokens near the beginning of the response receive a small bonus to encourage early formulation selection. 
An additional bonus is assigned to tokens corresponding to the structured field that specifies the selected OR expert, and a smaller bonus is applied to tokens associated with a nontrivial justification of sufficient length.

No token-level reward is given to long mathematical-model descriptions or code-like segments, which prevents the policy from being biased toward template over-generation.
After applying the response mask, the shaped token rewards are normalized within each rollout group sharing the same prompt.

\end{document}